\documentclass[conference]{IEEEtran}
\IEEEoverridecommandlockouts

\usepackage{times}
\usepackage[numbers,square,compress]{natbib}
\usepackage{graphicx}
\usepackage{hyperref}
\usepackage{float}
\usepackage{amsmath,amssymb}
\usepackage{booktabs}
\usepackage{multirow}
\usepackage{array}

\begin{document}

\title{A Dexterous and Compliant Gripper With Soft Hydraulic Actuation for Microgravity Manipulation}

\author{
William Su$^{1}$,
Jordan Kam$^{1}$,
Yixiao Wang$^{2}$,
Jianshu Zhou$^{3}$

\thanks{This work was funded by the Jacobs Institute for Design Innovation through the Innovation Catalysts Spark Grant and the National University of Singapore through the NUS Startup Grant (FY2026). Jordan Kam was supported by NASA's California Space Grant Consortium under Award \#80NSSP20M0099.}
\thanks{$^{1}$Aerospace Engineering Program, University of California, Berkeley, Berkeley, CA 94720, USA.}
\thanks{$^{2}$Department of Mechanical Engineering, University of California, Berkeley, Berkeley, CA 94720, USA.}
\thanks{$^{3}$Department of Mechanical Engineering, National University of Singapore, Singapore 117575.}
}

\maketitle

\begin{abstract}

Astrobee’s existing one-degree-of-freedom (DOF) underactuated compliant claw gripper enables perching on the International Space Station (ISS), but provides limited capability for continuous dexterous manipulation. More complex microgravity tasks require an end-effector that can maintain stable contact while limiting disturbance to the free-flying base, since contact forces directly couple into base motion. This article presents the integration of DexCoHand, a dexterous and compliant two-finger, 6-DOF gripper, with the Astrobee free-flying robot for microgravity manipulation. The system is evaluated in MuJoCo using Astrobee’s standard handrail perching sequence, including approach, perching, and subsequent pan and tilt motions. Compared with Astrobee’s existing gripper, DexCoHand preserves the commanded pan and tilt motions while reducing unintended cross-axis base motion. Hardware experiments on Earth further demonstrate DexCoHand’s dexterous manipulation capabilities and its potential for more adaptable intelligent manipulation tasks. 

\end{abstract}


\section{Introduction}

Astrobee is a free-flying robot aboard the International Space Station (ISS) that performs autonomous inspection, localization, docking, and perching tasks \cite{smith2016astrobee, oestreich2021orbit, soussan2022astroloc, park2017developing}. One key manipulation task Astrobee is capable of is perching onto handrails along the ISS Kibo module using its perching arm payload \cite{park2017developing}. Astrobee does so using a 1-DOF underactuated compliant claw gripper designed to perch around the ISS handrail. This design is effective for perching, but it constrains Astrobee to a limited grasp-and-release manipulation regime. More dexterous tasks, such as reorienting a payload for handoff, aligning a tool tip, threading or routing a cable, and manipulating small objects, require pose regulation within the grasp rather than repeated regrasping \cite{morton2025deformable, wang2025long, cutkosky1989grasp}. In microgravity, this distinction matters. Objects do not settle under gravity, and every contact impulse couples directly into the free-flyer base \cite{yoshida2003space}, inducing undesired translation or rotation and forcing compensatory control \cite{moosavian2007free}.

In recent years, several grippers have been developed to expand Astrobee's manipulation capabilities. Gecko-inspired adhesive grippers have been designed to enable anchoring on smooth surfaces \cite{chen2022testing}, compliant perching grippers provide stable handrail grasping \cite{park2017developing}, rigid-soft hybrid grippers improve conforming grasp of varied geometries \cite{kam2025underactuated}, and three-finger logistics grippers support cargo transfer and transport \cite{kam2025underactuated}. These grippers advance attachment, grasp robustness, and transport, but they provide limited support for continuous dexterous in-hand manipulation \cite{yousef2011tactile}.

\begin{figure}[t]
\centering
\includegraphics[width=\linewidth]{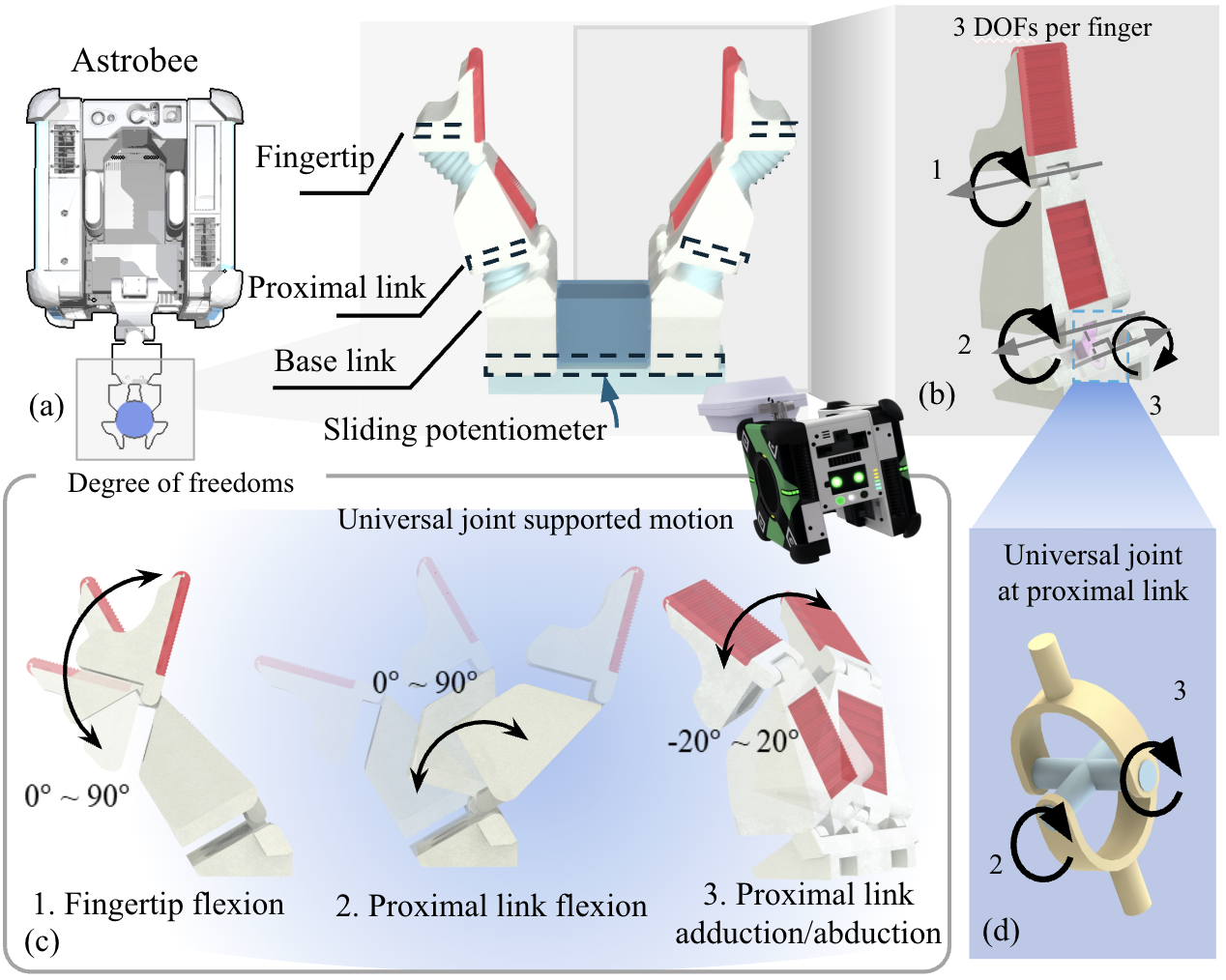}
\caption{(a) Astrobee with the DexCoHand attached on the perching arm. (b) 3-DOF per finger. (c) Supported finger motions, including fingertip flexion, proximal link flexion, and proximal link adduction/abduction. (d) Universal joint at the proximal link.}
\label{fig:system}
\end{figure}

\begin{figure*}[t]
\centering
\includegraphics[width=\linewidth]{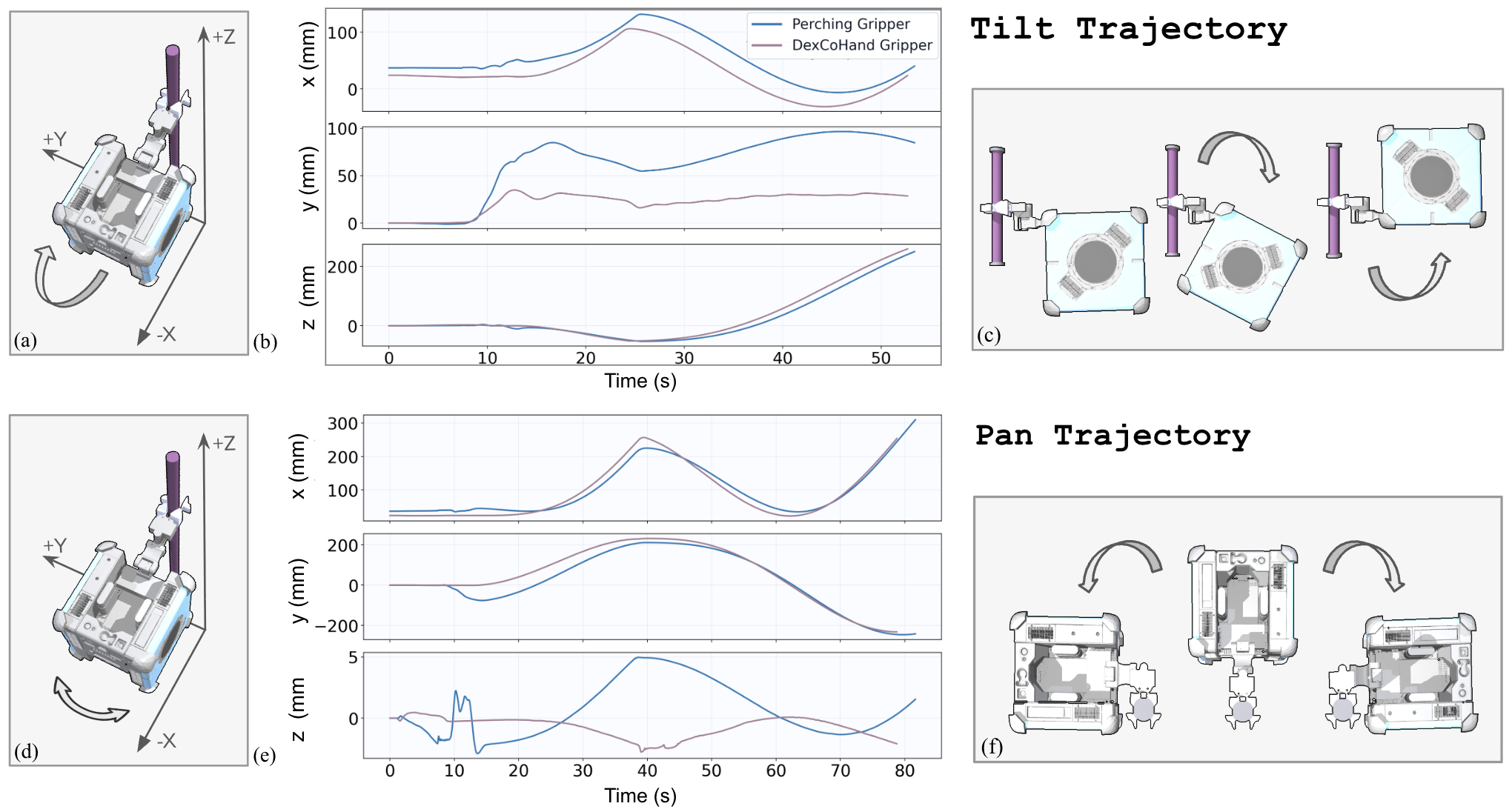}
\caption{Comparison of base motion during tilt and pan maneuvers with the perching gripper and DexCoHand. 
(a, d) tilt and pan maneuver setup and coordinate frame definition. 
(b, e) Base position trajectories along $x$, $y$, and $z$ during tilt and pan. 
(c, f) Representative tilt and pan motion sequence.}
\label{fig:results}
\end{figure*}

This article presents the integration of DexCoHand, a dexterous and compliant two-finger, six-degree-of-freedom (2F--6-DOF) gripper, with Astrobee's perching arm for microgravity manipulation \cite{zhou2024dexterous, zhou_plc}. Fig. \ref{fig:system} provides an overview of the integration of DexCoHand with Astrobee and the gripper's components and motion. DexCoHand introduces multiple controlled contact directions through 3-DOF fingers, while soft hydraulic actuation provides compliant force transmission under contact uncertainty \cite{zhou_antagonistic_pump}. We use DexCoHand to extend Astrobee’s dexterous manipulation capabilities from discrete grasp-and-release actions to continuous contact-rich manipulation while preserving the perching behavior that the current arm already performs. This increased dexterity would enable Astrobee to do more intelligent tasks in microgravity. 

\section{Methodology}

To evaluate DexCoHand's manipulation capabilities, we create two models: Astrobee with the DexCoHand and with the perching arm gripper \cite{nasa_astrobee_github}. We evaluate the DexCoHand gripper in MuJoCo compared to Astrobee's perching arm gripper using a standard handrail perching sequence consisting of approach, perching, and subsequent pan and tilt motions.  We compare base motion between the two end-effectors and use hardware demonstrations to show the dexterous manipulation capabilities enabled by DexCoHand. Together, these results show that the DexCoHand can perform the same manipulation tasks as the state-of-the-art perching arm gripper, but can also expand the manipulation space toward more intelligent dexterity in microgravity.

\subsection{Kinematics \& Contact Compliance}

DexCoHand provides two fingers with three controlled DOF per finger, including fingertip flexion, proximal flexion, and proximal adduction/abduction, as shown in Fig.~\ref{fig:system}. The flexion joints provide grasp closure, while the adduction/abduction joint give each finger lateral motion inside the grasp. For a contact point on the fingertip, the local contact velocity is given by:
\begin{equation}
\dot{\mathbf{x}}_c = \mathbf{J}_c(\mathbf{q})\dot{\mathbf{q}}
\label{eq:contact_jacobian}
\end{equation}
where $\mathbf{q}$ is the hand joint vector and $\mathbf{J}_c$ maps joint motion to contact motion. Increasing the number of independent finger DOF expands the column space of $\mathbf{J}_c$, allowing the contact point to move in additional directions on the object surface without requiring base motion or regrasping.

The soft hydraulic transmission introduces compliance between actuator command and contact force. Around a local operating point, this behavior can be represented using an effective compliance model. If $\mathbf{C}_q$ denotes joint-space compliance, the contact-frame compliance can be written as:
\begin{equation}
\mathbf{C}_c = \mathbf{J}_c \mathbf{C}_q \mathbf{J}_c^\top 
\label{eq:contact_compliance}
\end{equation}
This relationship allows compliance at the joints which maps directly to compliance at the contact interface. A more compliant contact reduces abrupt force changes during stick-slip transitions and distributes interaction forces over time, which helps limit unintended base motion during manipulation.

\subsection{Floating-Base Dynamics Model}

Astrobee cannot rely on a fixed ground reaction during manipulation. Contact forces at the gripper therefore appear directly as disturbances on the free-flying base. A simplified translational model used in simulation is:

\begin{equation}
m_b \dot{\mathbf{v}}_b =
\mathbf{F}_{\mathrm{thr}} - \sum_i \mathbf{f}_{c,i},
\label{eq:base_dynamics}
\end{equation}

where $m_b$ is the Astrobee base mass, $\mathbf{F}_{\mathrm{thr}}$ is the commanded propulsion force, and $\mathbf{f}_{c,i}$ are contact forces transmitted through the end-effector. In microgravity, gravity does not dissipate contact events, so the timing and direction of $\mathbf{f}_{c,i}$ directly affect base motion. We therefore log Astrobee base position during pan and tilt maneuvers as a proxy for contact-induced disturbance. Motion along the commanded direction indicates whether the perching maneuver is preserved, while motion in the orthogonal axes reflects coupling from the gripper into the free-flying base. This simplified model captures the net effect of contact forces, while full free-flying dynamics arise from coupled inertia and momentum exchange between the manipulator and spacecraft \cite{yoshida2003space}.

\subsection{Simulation Setup}

Astrobee with both gripper payloads is modeled in MuJoCo as a free-floating rigid body with grippers attached to the perching arm's end joint. The handrail is represented as a fixed cylindrical contact object. The simulation is performed in zero gravity so that the measured base motion comes from commanded actuation and contact interaction rather than conditions that would be seen on-Earth \cite{kam2025towards}. We evaluate both grippers under the same task sequence. While the gripper geometry and contact behavior differ, the perching task, rail location, and commanded arm motions are the same. This evaluation approach allows us to isolate the effect of the gripper at the end of the arm on contact-induced base motion. We log the Astrobee base position in the world frame during the pan and tilt motions. The logged trajectories are the translational base motion along the $x$, $y$, and $z$ axes, expressed in millimeters (mm).

\subsection{Evaluation}

Each trial follows the same rail-perching sequence of approaching, perching, and executing a tilt or pan maneuver. The tilt maneuver rotates Astrobee along the rail plane, while the pan maneuver sweeps laterally about the rail contact. This comparison highlights how the choice of gripper affects the motion of the Astrobee base during these controlled maneuvers. Motion along the commanded direction indicates whether the intended maneuver is preserved, while motion in the orthogonal axes reflects contact-induced coupling into the free-flyer. This formulation evaluates how the gripper affects base motion by comparing whether the commanded maneuver is preserved while limiting unintended motion in the orthogonal axes.

\section{Results}

Both grippers successfully execute pan and tilt motions under contact constraints, with the primary motion axes preserved across both models. However, differences emerge in cross-axis motion. During tilt, the perching gripper exhibits increased lateral deviation along the orthogonal $y$ axis, reaching approximately 80\,mm, while DexCoHand maintains significantly lower deviation at approximately 20\,mm, as shown in Fig.~\ref{fig:results}(b). Similarly, during pan, DexCoHand remains near zero deviation, approximately in the range of 0 to $-2$\,mm, whereas the perching gripper shows larger variation, ranging from $-2$ to 4\,mm, as shown in Fig.~\ref{fig:results}(e). These results indicate that DexCoHand maintains nominal motion execution with comparably small unintended motion in orthogonal directions, which is a necessary condition for deploying dexterous grippers on free-flyers. Additional hardware experiments further demonstrate capabilities not achievable with compliant claw-like grasping grippers, including in-hand object reorientation, controlled pivoting, and compliant pinching \cite{zhou_unified_manipulability}.

\begin{figure}[t]
\centering
\includegraphics[width=\linewidth]{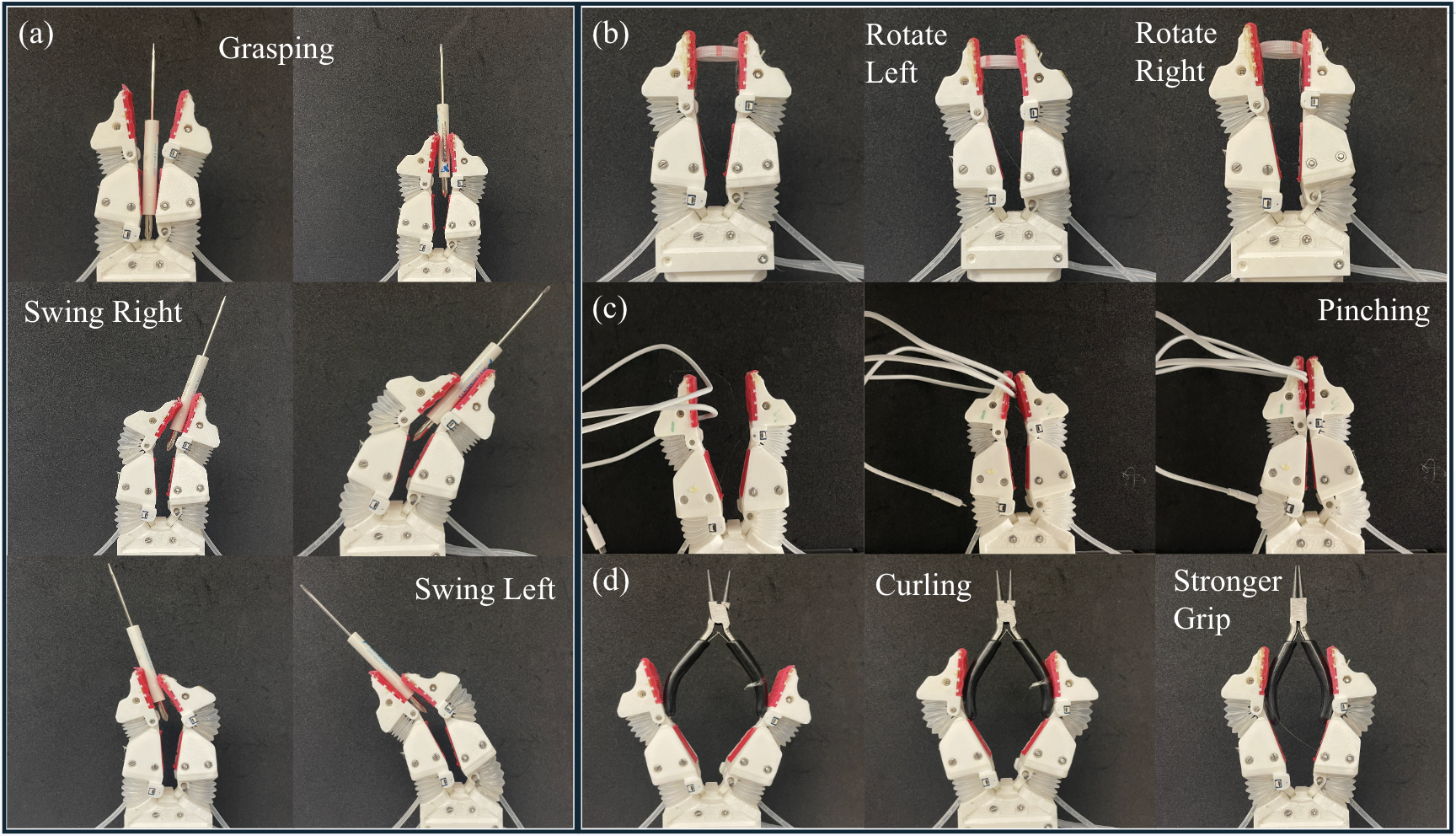}
\caption{Dexterous manipulation capabilities demonstrated by DexCoHand including (a) Grasping and swinging motions. 
(b) In-hand object reorientation. 
(c) Pinching using compliant fingertip contact.  (d) Conformal contact on irregular geometries.}
\label{fig:tasks}
\end{figure}

\section{Space Applications}

The ability to perform continuous in-hand manipulation enables increased dexterous tasks in space. Such tasks include tool alignment, cable routing \cite{wang2025long}, manipulation of flexible payloads \cite{morton2025deformable}, and interaction with uncertain or deformable objects \cite{moreira2020cooperative}. Unlike rigid grasping strategies, compliant multi-DOF manipulation allows the system to regulate contact forces while maintaining stability. This is particularly important in microgravity, where disturbances propagate directly to the base system. Fig.~\ref{fig:tasks} presents a conceptual overview of the DexCoHand manipulating objects that could be of interest for free-flying manipulation \cite{zhou_soft_syringes}. While this work has demonstrated DexCoHand's manipulation capabilities on Astrobee, the manipulation architecture generalizes to other free-flyers such as the Synchronized Position Hold, Engage, Reorient, Experimental Satellites (SPHERES) free-flyer \cite{miller2000spheres, estrada2017force}. The integration of dexterous, compliant grippers provides a pathway toward more autonomous and adaptable robotic operations in microgravity.

\section{Conclusion}

This article presents the integration of the DexCoHand, a dexterous compliant gripper, with a free-flying robot. The results demonstrate that DexCoHand preserves baseline perching capabilities while enabling new dexterous manipulation tasks. The key contribution is a conceptual validation that dexterous manipulation can be introduced into free-flyers without compromising stability. Future work will extend this framework to full in-hand manipulation tasks under closed-loop control in simulation with the same objects outlined in the applications section of this article.

\newpage

\bibliographystyle{IEEEtranN}
\bibliography{references}

@inproceedings{oestreich2021orbit,
  title={On-orbit inspection of an unknown, tumbling target using NASA's Astrobee robotic free-flyers},
  author={Oestreich, Charles and Espinoza, Antonio Ter{\'a}n and Todd, Jessica and Albee, Keenan and Linares, Richard},
  booktitle={Proceedings of the IEEE/CVF Conference on Computer Vision and Pattern Recognition},
  pages={2039--2047},
  year={2021}
}

@inproceedings{soussan2022astroloc,
  title={Astroloc: An efficient and robust localizer for a free-flying robot},
  author={Soussan, Ryan and Kumar, Varsha and Coltin, Brian and Smith, Trey},
  booktitle={2022 International Conference on Robotics and Automation (ICRA)},
  pages={4106--4112},
  year={2022},
  organization={IEEE}
}

@article{cutkosky1989grasp,
  title={On grasp choice, grasp models, and the design of hands for manufacturing tasks.},
  author={Cutkosky, Mark R and others},
  journal={IEEE Transactions on robotics and automation},
  volume={5},
  number={3},
  pages={269--279},
  year={1989}
}

@inproceedings{wang2025long,
  title={Long-Reach Robotic Manipulation for Assembly and Outfitting of Lunar Structures},
  author={Wang, Stanley and Kojouharov, Venny and Chung, Long Yin and Morton, Daniel and Cutkosky, Mark},
  booktitle={2025 International Conference on Space Robotics (iSpaRo)},
  pages={498--504},
  year={2025},
  organization={IEEE}
}

@article{moosavian2007free,
  title={Free-flying robots in space: an overview of dynamics modeling, planning and control},
  author={Moosavian, S Ali A and Papadopoulos, Evangelos},
  journal={Robotica},
  volume={25},
  number={5},
  pages={537--547},
  year={2007},
  publisher={Cambridge University Press}
}

@article{morton2025deformable,
  title={Deformable Cargo Transport in Microgravity with Astrobee},
  author={Morton, Daniel and Antonova, Rika and Coltin, Brian and Pavone, Marco and Bohg, Jeannette},
  journal={arXiv preprint arXiv:2505.01630},
  year={2025}
}

@inproceedings{park2017developing,
  title={Developing a 3-DOF compliant perching arm for a free-flying robot on the International Space Station},
  author={Park, In-Won and Smith, Trey},
  booktitle={IEEE International Conference on Advanced Intelligent Mechatronics (AIM)},
  year={2017}
}

@inproceedings{moreira2020cooperative,
  title={Cooperative Real-Time Inertial Parameter Estimation},
  author={Moreira, Marina and Coltin, Brian and Ventura, Rodrigo},
  booktitle={19th International Conference on Autonomous Agents and MultiAgent Systems},
  year={2020}
}

@article{zhou2024dexterous,
  title={A dexterous and compliant (DexCo) hand based on soft hydraulic actuation for human-inspired fine in-hand manipulation},
  author={Zhou, Jianshu and Huang, Junda and Dou, Qi and Abbeel, Pieter and Liu, Yunhui},
  journal={IEEE Transactions on Robotics},
  volume={41},
  pages={666--686},
  year={2024},
  publisher={IEEE}
}

@article{zhou_unified_manipulability,
  author    = {Zhou, J. and Liang, B. and Huang, J. and Tomizuka, M. (2025)},
  title     = {Unified Manipulability and Compliance Analysis of Modular Soft-Rigid Hybrid Fingers},
  journal   = {IFAC-PapersOnLine, 59(30), 335-340},
  year      = {2025}
}

@article{zhou_soft_syringes,
  author    = {Jianshu Zhou and Junda Huang and X. Ma and A. Lee and K. Kosuge and Y. H. Liu},
  title     = {Design, Modeling, and Control of Soft Syringes Enabling Two Pumping Modes for Pneumatic Robot Applications},
  journal   = {IEEE/ASME Transactions on Mechatronics},
  year      = {2024},
  volume    = {29},
  number    = {2},
  pages     = {889--901}
}

@inproceedings{kam2025underactuated,
  title={An underactuated rigid-soft hybrid gripper prototype for conforming free-flying manipulation},
  author={Kam, Jordan and Vargas, Andres Mora and Coltin, Brian},
  booktitle={IEEE International Conference on Robotics and Automation},
  year={2025}
}

@article{yousef2011tactile,
  title={Tactile sensing for dexterous in-hand manipulation in robotics—A review},
  author={Yousef, Hanna and Boukallel, Mehdi and Althoefer, Kaspar},
  journal={Sensors and Actuators A: physical},
  volume={167},
  number={2},
  pages={171--187},
  year={2011},
  publisher={Elsevier}
}

@misc{nasa_astrobee_github,
  author = {{NASA}},
  title = {Astrobee Robot Software},
  howpublished = {\url{https://github.com/nasa/astrobee}}
}

@inproceedings{kam2025towards,
  title={Towards a microgravity sim-to-real training environment for robotic systems in low earth orbit},
  author={Kam, Jordan and Darrell, Trevor},
  booktitle={2025 Regional Student Conferences},
  pages={97394},
  year={2025}
}

@inproceedings{estrada2017force,
  title={Force and moment constraints of a curved surface gripper and wrist for assistive free flyers},
  author={Estrada, Matthew A and Jiang, Hao and Noll, Bessie and Hawkes, Elliot W and Pavone, Marco and Cutkosky, Mark R},
  booktitle={2017 IEEE International Conference on Robotics and Automation (ICRA)},
  pages={2824--2830},
  year={2017},
  organization={IEEE}
}

@inproceedings{miller2000spheres,
  title={SPHERES: a testbed for long duration satellite formation flying in micro-gravity conditions},
  author={Miller, David and Saenz-Otero, A and Wertz, J and Chen, A and Berkowski, G and Brodel, C and Carlson, S and Carpenter, D and Chen, S and Cheng, S and others},
  booktitle={Proceedings of the AAS/AIAA space flight mechanics meeting},
  volume={105},
  pages={167--179},
  year={2000},
  organization={Clearwater, Florida, January}
}

@article{chen2022testing,
  title={Testing gecko-inspired adhesives with astrobee aboard the international space station: Readying the technology for space},
  author={Chen, Tony G and Cauligi, Abhishek and Suresh, Srinivasan A and Pavone, Marco and Cutkosky, Mark R},
  journal={IEEE Robotics \& Automation Magazine},
  volume={29},
  number={3},
  pages={24--33},
  year={2022},
  publisher={IEEE}
}

@article{zhou_antagonistic_pump,
  author    = {Jianshu Zhou and W. Chen and H. Cao and Q. He and Y. Liu},
  title     = {Antagonistic Pump with Multiple Pumping Modes for On-Demand Soft Robot Actuation and Control},
  journal   = {IEEE/ASME Transactions on Mechatronics},
  year      = {2023},
  volume    = {29},
  number    = {5},
  pages     = {3252--3264}
}

@article{zhou_plc,
  author    = {J. Zhou and W. Chen and J. Huang and B. Liang and Y. Liu and M. Tomizuka},
  title     = {Programmable Locking Cells (PLC) for Modular Robots with High Stiffness Tunability and Morphological Adaptability},
  journal   = {IEEE Transactions on Robotics},
  year      = {2025}
}

@article{smith2016astrobee,
  author = {Trey Smith and Mark Bualat and Terry Fong and Jeremy Smith and Ernest Wong and others},
  title = {Astrobee: A New Platform for Free-Flying Robotics Research on the International Space Station},
  journal = {Journal of Field Robotics},
  year = {2016}
}

@article{yoshida2003space,
  author = {Kazuya Yoshida},
  title = {Engineering Test Satellite VII Flight Experiments For Space Robot Dynamics and Control: Theories on Laboratory Test Beds Ten Years Ago, Now in Orbit},
  journal = {The International Journal of Robotics Research},
  volume = {22},
  number = {5},
  pages = {321--335},
  year = {2003}
}

\end{document}